\documentclass[10pt,twocolumn,letterpaper]{article}

\usepackage{cvpr}
\usepackage{times}
\usepackage{epsfig}
\usepackage{graphicx}
\usepackage{subcaption}
\usepackage{multirow}
\usepackage{authblk}
\usepackage{dblfloatfix}
\usepackage[linesnumbered,ruled,vlined]{algorithm2e}
\SetKwInput{Input}{Input}
\SetKwInput{Output}{Output}
\SetKwInput{Param}{Param}

\usepackage{amsmath}
\usepackage{amssymb}
\usepackage[toc,page]{appendix}

\usepackage[pagebackref=true,breaklinks=true,letterpaper=true,colorlinks,bookmarks=false]{hyperref}

\cvprfinalcopy 

\def\cvprPaperID{9580} 

\begin{document}
\title{The Knowledge Within: Methods for Data-Free Model Compression}

\author[1,2]{Matan Haroush, Itay Hubara}
\author[1]{Elad Hoffer}
\author[2]{Daniel Soudry}
\small\affil[1]{Habana Labs Research, Caesarea, Israel}
\small\affil[2]{Department of Electrical Engineering, Technion, Haifa , Israel\\
\tt\small{\{mharoush,ihubara,ehoffer\}@habana.ai},
\tt\small{daniel.soudry@gmail.com}
}

\maketitle
\thispagestyle{empty}
\begin{abstract}
\textbf{Background:} Recently, an extensive amount of research has been focused on compressing and accelerating Deep Neural Networks (DNN). So far, high compression rate algorithms require part of the training dataset for a low precision calibration, or a fine-tuning process. However, this requirement is unacceptable when the data is unavailable or contains sensitive information, as in medical and biometric use-cases.

\textbf{Contributions:} We present three methods for generating synthetic samples from trained models. Then, we demonstrate how these samples can be used to calibrate and fine-tune quantized models without using any real data in the process. Our best performing method has a negligible accuracy degradation compared to the original training set. This method, which leverages intrinsic batch normalization layers' statistics of the trained model, can be used to evaluate data similarity. Our approach opens a path towards genuine data-free model compression, alleviating the need for training data during model deployment.

\date{\vspace{-3ex}}
\end{abstract}
\begin{figure}[t]
    \includegraphics[width=0.47\textwidth,angle=270]{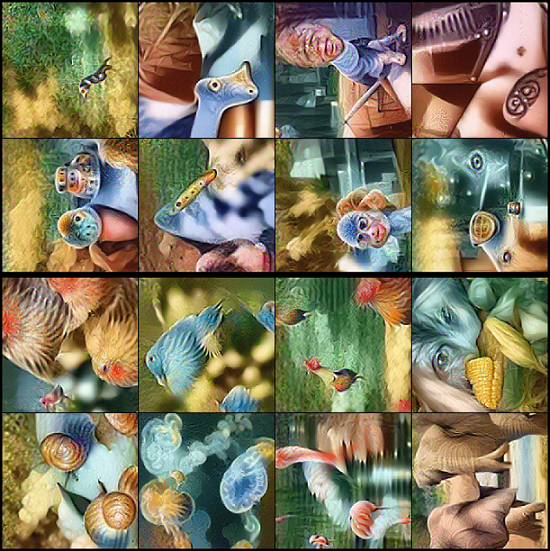}
    \caption{Synthetic samples generated from ResNet-18 trained on Imagenet, using our methods: $BNS$ (right;\ref{sec:bns}) and the class-conditional $BNS+\mathcal{I}$ (left;\ref{sec:bns+i}). Additional samples are available in appendix-\ref{apdx:gen_samples}.}
      \label{fig:visual_demo1}
      \date{\vspace{-3ex}}
\end{figure}
\section{Introduction}
Quantization is a prevalent, accelerator friendly, compression method \cite{hubara2017quantized,jacob2018quantization} employed prior to DNNs deployment within real-world applications. However, high compression rates typically demand additional information to minimize quality loss. For example, when using low precision arithmetic, it is often beneficial to trade-off the numerical representation range with its resolution by clipping the dynamic range based on its real expected values. Thus, it is a common practice to gather per tensor statistics \cite{jacob2018quantization,banner2019post} from a subset of samples drawn from the training set (Calibration). Furthermore, in cases of very high compression rates, model parameters often require additional adjustment to recover from catastrophic error accumulation, where access to the entire training data is needed.
This may lead to an undesired coupling between deployment and training phases of a model through data. Especially in cases where the training data is sensitive or simply unavailable at the time of deployment. Therefore, it is appealing to investigate new methods to alleviate the need for real data for deployment purposes, for example, via substitution with synthetic data.

Generating high-quality samples requires capturing the prior distribution of the data which is often hard. A large body of work dedicated to generative models has shown that it is possible to learn such priors and generate high-resolution synthetic images \cite{kingma2013autoencoding,goodfellow2014generative,salimans2016improved, heusel2017gans,  brock2018large}. Unfortunately, these techniques come at the cost of training a dedicated generative model. This training requires access to the real training data which we want to avoid.

We aim to understand the potential and limitations of synthetic samples for model compression tasks, specifically for reduced precision deployment i.e., calibration and fine-tune via Knowledge Distillation (KD \cite{hinton2015distilling}). In the context of this work, the full precision model (or parts of it) serves as the teacher, while the student is the low precision counterpart. We will focus our attention on DNNs for classification tasks (e.g. CIFAR and ImageNet \cite{krizhevsky2009learning,russakovsky2015imagenet}). In the process, we revisit a feature visualization process dubbed "Inceptionism" \cite{mordvintsev2015inceptionism} which is based solely on the trained model. This process typically starts with an arbitrary input, which is iteratively adjusted to maximize the response of a set of target features via back-propagation. Furthermore, the optimization process is typically constrained using some prior knowledge about the input, such as a high correlation between nearby pixels within an image, to avoid over-fitting the generated sample. A similar approach was recently adapted for several use-cases including for data-free distillation \cite{lopes2017data, bhardwaj2019dream,nayak2019zero} with limited success.

Our contribution is twofold: First, we offer novel methods for generating and leveraging synthetic samples for the use of knowledge distillation. These samples are created under a realistic data-free regime by exploiting the encapsulated knowledge within the provided model. We then empirically evaluate the usefulness of these samples for model compression, yielding comparable results to the original training dataset with minimal accuracy degradation. Second, we propose a novel approach for evaluating the similarity between a reference dataset and a set of arbitrary samples. In a nutshell, we suggest measuring the mean divergence of the second-order statistics drawn from a set of intermediate layers of a given model that was trained on the reference dataset (reference model). This can be done without relying on access to any real data by leveraging intrinsic measurements provided by Batch Normalization (BN) layers \cite{ioffe2015batch}.
\section{Related work}
The notion of "data-free" quantization was recently introduced by Nagel et al., \cite{nagel2019data} who proposed using the pre-determined measurements of batch-normalization statistics \cite{ioffe2015batch}, that are gathered during training to determine the proper dynamic range for each layer. However, this method requires either access to all of the layers' statistics or relying on a closed-form analytical solution for layers for which measurements are not available, based on their input distribution. Furthermore, this method is unable to handle cases where calibration leads to extreme degradation, so fine-tuning is required.

Previous work explored using data-free knowledge distillation for model compression by collecting metadata related to the statistics of a trained model output \cite{lopes2017data,bhardwaj2019dream}. In both cases, the generation scheme consists of sampling a set of random target tensors and minimizing the mean square error between the output and the sampled targets. Lopes et al.,\cite{lopes2017data} proposed collecting means and co-variance for a set of layers after training, while optimization targets are sampled per layer independently under multivariate Gaussian assumption. The authors mention this method fails to capture inter-layer relations and proposed an alternative approach to capture inter-layer relations via graph spectral analysis with compelling results for models trained on the MNIST dataset. However, this is impractical for large models or models with large input size, due to its computational cost.

Bhardwaj et al., \cite{bhardwaj2019dream} suggested it is sufficient to collect metadata from the layer before the linear classifier. Metadata collection is done by processing a small portion of the training set, clustering the high dimensional outputs and applying Principle Component Analysis (PCA) per cluster. The optimization target is based on the collected centroids and a random noise projected in the direction of the primary principle components. Bhardwaj et al., \cite{bhardwaj2019dream} presented a small set of experiments on CIFAR10 dataset, with relatively large degradation between real data to the generated samples.

Recently, Nayak et al., \cite{nayak2019zero} proposed similar data-free method dubbed as zero-shot distillation, which relies solely on the final layer weights to compute a class similarity matrix. Then, sample soft targets via Dirichlet distribution for generating synthetic images. However, KD results on synthetic samples are only comparable to real data on easily separable datasets such as MNIST, while performing poorly on CIFAR10.

In this work, we aim to generate samples that mimic the training data distribution by some measure. This can be achieved with naive noise sampling according to the low-order statistics of original data, or by directly optimizing a similarity measure of the internal activations' statistics in a trained model. In contrast to previous attempts \cite{lopes2017data,bhardwaj2019dream} which required sampling activation generated by real data, our method relies on low order statistics captured by existing BN layers (i.e., channel-wise mean and standard deviation). We then track the statistics induced by the synthetic samples and directly optimize a divergence score with respect to the reference set, while relying on the model structure to naturally maintain inter-layer statistics' relations.

\section{Methods for data-free distillation}
\label{gen_methods}
We are interested in a data-free regime, wherein a model is given without its corresponding dataset used for training. This regime reflects a realistic scenario, as training data is often confidential or private. Therefore, we offer three methods for generating useful synthetic data for distillation and calibration:
\begin{itemize}
    \item\textbf{Gaussian scheme:} samples are randomly drawn from a Gaussian distribution.
    \item\textbf{Inception scheme:} samples are generated via logit maximization (i.e., a special case of the Inceptionism scheme).
    \item\textbf{BN-Statistics scheme:} samples are generated by optimizing a novel internal statistics' divergence measure.
\end{itemize}
\subsection{Gaussian Scheme}
As a naive approximation of the original dataset, it is natural to consider a simple Gaussian generator (denoted as $\mathcal{G}$) with first and second moments defined to match the original input data. We suggest this alone may be sufficient for model calibration tasks required for low numerical precision inference, under mild compression demands. Such a generation scheme is appealing since an indefinite number of samples can be generated at will with minimal compute and storage requirements. However, as one would expect, under extreme compression demands, when the model parameters need to be adjusted (e.g., fine-tune using distillation), this method may prove to be insufficient.

Since this method does not preserve the original input's structure, the internal activation's statistics may differ significantly from the original statistic induced by the training data. This change can harm the model accuracy especially if it contains Batch-Normalization (BN) layers \cite{ioffe2015batch}. BN layers are commonly used in DNNs workloads, as they have been shown to improve both accuracy and speed of convergence by normalizing the input to the layer to have zero mean and unit variance before applying its operation. During training, each BN layer keeps a running estimate of the empirical mean and standard-deviation per-channel of its inputs which latter applied to normalized the input data during inference. 

Adjusting the model's parameters by using distillation over the randomly generated samples, will irretrievably alter the existing BN layer parameter to accommodate for the observed statistical properties, resulting in an imminent failure when turning back to evaluate real data. Thus, we suggest forcing all BN layers in the model to maintain their original running estimates for evaluation. As shown in section \ref{experimnets} this may negate this obstacle to some extent, enabling the use of such samples for the task of distillation.

\subsection{Inception Scheme}
Inceptionism \cite{mordvintsev2015inceptionism} related generation schemes, typically impose constraints solely on the input and output of the model. We shall focus on a special case, which we term the Inception scheme (denoted as $\mathcal{I}$), where only a single neuron from the final model output is maximized under an input smoothness requirement.
We define the optimization objective is the sum of a Domain Prior term and an Inception loss term. As a domain prior, we use a Gaussian smoothing kernel to produce a smoothed version of the provided input. The loss term is then computed as the mean squared error between the input image and the smoothed variant, encouraging nearby pixels to have similar values. The Inception Loss term is derived by choosing an arbitrary label and perform gradient descent on the negative exponent of the appropriate class logit drawn from the Fully Connected (FC) layer output, i.e. $e^{-logit/scale}$. The Inception Loss injects the desired class information, where the exponent and temperature-scale control the impact of the logit magnitude on the loss, preventing the model from producing inputs that cause the output to explode by exponentially decaying the logit contribution to the general loss as its magnitude increases. This scheme often results in high confidence outputs, however, it is highly sensitive to the hyper-parameters (see appendix-\ref{apdx:hp}).

\subsection{BN-Statistics Scheme}
\label{sec:bns}
Based on our Gaussian and Inception scheme experiments we hypothesize that the lack of regularization on the internal statistic during the sample generation process may result in significant internal statistics' divergence from the observed statistics on real data. Indicating such samples are not drawn from a distribution similar to the original data (as we show later, in Figure \ref{fig:inception_gen}). This, in turn, may impede their use for model compression and KD. Thus, our proposed approach is to directly minimize the internal statistics' divergence, by optimizing a novel measure which we term "BN-Stats" (denoted as $BNS$). $BNS$ only assumes access to the predetermined empirical measurements that occur for each BN layer. We will use these estimates as targets and compare new data samples by the similarity between their measured statistics.

More formally, given the running estimates $\hat{\mu}$ and $\hat{\sigma}$ of a given batch-norm layer, we wish to measure $\Tilde{\mu}=\mu(D)$ and $\Tilde{\sigma}=\sigma(D)$ of our new data activations $D$ and evaluate their similarity to the reference statistics. $BNS$ define this similarity using the Kullback-Leibler (KL) divergence under a simplistic isotropic Gaussian assumption such that 
\begin{align}
    \smaller
    \begin{split}
    BNS(D,\hat{\mu},\hat{\sigma}) &= KL\left(\mathcal{N}(\hat{\mu},\hat{\sigma}^2)||\mathcal{N}(\Tilde{\mu},\Tilde{\sigma}^2)\right) \\
    &= \log\frac{\Tilde{\sigma}}{\hat{\sigma}} - \frac{1}{2} \left(1 - \frac{\hat{\sigma}^2+(\hat{\mu}-\Tilde{\mu})^2}{\Tilde{\sigma}^2} \right)
    \end{split}
\end{align}
Our generation process starts with a batch of random input samples, which are then iterativly adjusted to minimize the mean statistical divergence across all layers in the reference set. Namely, for each optimization step over an input batch $X$, we extract per-BN-layer activation set $\{D_l\}_{l=1}^N$. We then define the optimization objective as the mean statistics' divergence over all $BNS(D_l,\hat\mu_l,\hat\sigma_l)$. Formally, we define the first and second moments sets as:
\begin{center}\smaller
$T,S:=\{P_l\sim\mathcal{N}(\hat{\mu}_l,\hat{\sigma}_l)\}_{l=1}^N,\{Q_l\sim\mathcal{N}(\Tilde{\mu}_l,\Tilde{\sigma}_l)\}_{l=1}^N$
\end{center}
while the optimization objective (i.e., BNS measure) is defined as the mean KL divergence on the set $\{T,S\}$, i.e. 
\begin{align}\smaller
\mathcal{J}_{KL}(X|T,S) := \frac{1}{N}\sum_{l=1}^NKL(P_l||Q_l) \\= \frac{1}{N}\sum_{l=1}^NBNS(D_l,\hat\mu_l,\hat\sigma_l)
\end{align}
A small $\epsilon=1e^{-8}$, is added to the measured variance of the induced synthetic distribution to accommodate for zero-variance channels. We note that alternative metrics can be used for $BNS$ under the same underlying assumptions, such as Mean Square Error over the empirical moments which is symmetric and can handle zero variance channels. Finally, standard backpropagation is used to adjust $X$.
\subsection{Combining $BNS+\mathcal{I}$}
\label{sec:bns+i}
In addition to the three schemes described above, we experiment with a combination of BN-Statistics and Inception schemes objectives denoted as $BNS+\mathcal{I}$, see Algorithm \ref{alg:bns_i_gen}. This method of generation attempts to harness the strength of each method. The class constraint imposed by $\mathcal{I}$ impacts the induced statistics depending on the batch composition, forcing the optimization process to compensate for the sampled classes. The drawback of this method is the additional loss scaling parameters which are now added to the original hyper-parameters of the Inception scheme. We didn't exhaustively investigate the best method to combine the two however, results of a simple aggregation provides an encouraging improvement in some cases.
\begin{algorithm}[h]\smaller\DontPrintSemicolon
    \Input{a pre-trained model with $N$ BN layers}
    \Param{$input\_shape, batch\_size,budget$}
    \Param{$class\_temp,\theta_{prior},\alpha,\beta,\gamma$}
    \Param{$optimizer(\theta),\#duplicates$}  
    \Output{a batch of synthetic samples | $X_{budget}$}
    Init: $X_0 \xleftarrow[]{} randn(batch\_size,input\_shape)$\\
    Init: $targets \xleftarrow[]{} rand\_int(batch\_size,\#classes)$ \\
    Extract $T:=\{\hat\mu_l,\hat\sigma_l\}_{l=1}^N$ \tcp*{\smaller $P_l\sim\mathcal{N}(\hat\mu_l,\hat\sigma_l)$ }
    \While{$i < budget$}
    {
        Compute: $X_i \xleftarrow[]{} clamp(X_i,0,1)$ \\
        Compute: $\hat{X}_i  \xleftarrow[]{} augment(X_i,\#duplicates)$ \\
        Compute: $loss_{p}  \xleftarrow[]{} prior(\hat{X}_i|\theta_{prior})$ \tcp*{\smaller e.g. $smooth(\hat{X}_i|k,\sigma)$}
        Compute: $logits  \xleftarrow[]{} forward(\hat{X}_i)$ \\
        Record:  $S:=\{\Tilde{\mu}_l,\Tilde{\sigma}_l\}_{l=1}^N$ \tcp*{\smaller $Q_l\sim\mathcal{N}(\Tilde{\mu}_l,\Tilde{\sigma}_l)$}
        Compute: $loss_{i}  \xleftarrow[]{} e^{-\frac{logits[targets]}{scale}}$ \\
        Compute: $loss_{s} \xleftarrow[]{} \mathcal{J}_{KL}(\hat{X}_i|T,S)$ \\
        Compute: $loss  \xleftarrow[]{} \alpha*loss_{s} + \beta*loss_{i} + \gamma*loss_{p}$ \\
        Compute: $G_{X_i}  \xleftarrow[]{} backward(loss,X_i)$ \tcp*{\smaller i.e., $\frac{\partial loss}{\partial{ X_i}}$}
        Update: $X_{i+1} \xleftarrow[]{} optimizer.step(G_{X_i},\theta)$  \\
        Update: $i \xleftarrow[]{} i+1 $ \\
    }
    
\caption{Generating $BNS+\mathcal{I}$ samples}
\label{alg:bns_i_gen}

\end{algorithm}
\section{Experiments}
\label{experimnets}
\subsection{Generating data samples}
\label{sec:gen_data_setting}

We first describe several components used in our sample generation methods. Specifically, note that $\mathcal{I}$ \& $BNS$ can be seen as special cases of the $BNS+\mathcal{I}$ method, by scaling the appropriate loss components. $BNS+\mathcal{I}$ method is described in algorithm-\ref{alg:bns_i_gen}, the reader may refer to appendix-\ref{apdx:hp} for specific hyperparameters details.

\begin{itemize}
\item\textbf{Inception Loss:} The Inception loss is defined as the exponent of the negative value of the appropriate class logit drawn from the Fully Connected (FC) layer of the model i.e., $e^{-logit/scale}$. A random set of labels are chosen as targets.

\item\textbf{Image Prior:} Nearby pixels of an image often have similar values, common techniques attempt to encourage such relations. In practice we use a Gaussian smoothing prior, by applying a convolution with a Gaussian kernel on the input, creating a smoothed version that is then used to compute the mean squared error from the original input.

\item\textbf{Statistics Loss:} we define $\mathcal{J}_{KL}$ as our statistics loss. Reference statistics are extracted from all BN layers in the model. Those are simply the running mean and variance gathered during training. Additionally, we treat the reported dataset normalization in the same manner.

\item \textbf{Input Trimming:} At the start of each optimization step we clip the input values to $[0,1]$ range to match real data values.
\item\textbf{In-Batch Augmentations:} Since activation gradients are not aggregated across batch during back standard propagation, we use in-batch augmentations \cite{hoffer2019augment} technique to gain a gradient smoothing effect per sample; Specifically, each sample is duplicated $N$ times with a set of random differential augmentations chosen from a set containing Random-N-Cutout, Crop-Resize and Flip.
\end{itemize}
Since the generation process must start with a previously trained model, we concentrate most of our experiments on ResNet meta-architecture \cite{he2016deep, Zagoruyko_2016}, as it is a popular example of an architecture in which training quality heavily relies on batch normalization layers.

\subsection{Data-free model compression}
\textbf{General settings.} Model compression for deployment usually requires some additional effort. The compression level correlates with the amount of information and work necessary to reduce accuracy degradation. Using low numerical precision as a model compression technique for inference typically requires a prior calibration step to determine the dynamic range (which sets the scale and zero-point values) \cite{krishnamoorthi2018quantizing} for each intermediate layer. This restriction sacrifices the dynamic range for an improved numerical resolution. When enforcing highly demanding compression rates on a model it is often required to retrain the model under similar compression constraints to recover from any significant accuracy loss. The process of adjusting the model parameters to the compression constrains is commonly called fine-tuning. Since each model responds differently to quantization, we choose numerical precision to ensure that calibration or fine-tuning is needed. We will consistently use the notation of $\#w\#a$, to denote the number of numerical precision bits used for representing the quantized activations and weights throughout the model.

In all our experiments the case of uniform quantization is considered, with per-channel scale for the weights and a per-tensor scale for the layer's activation values as detailed in \cite{krishnamoorthi2018quantizing}. A copy of the weights \cite{hubara2017quantized}, as well as the biases and gradients, are kept in single-precision where the latter is derived using a straight-through estimator \cite{bengio2013estimating}. All results are reported for simulated quantization (i.e. discretizing the inputs and weights before applying float operators). Additionally, Batch-Normalization layers were kept in single precision.

\textbf{Calibration details.}\label{sec:calib} Our experiments begin by calibrating each model using real and synthetic data, under increasing compression rates. During calibration, we use a smoothed absolute dynamic range measurement. That is, the dynamic range is measured via running estimates of the mean absolute min/max values within chunks containing 16 samples each. We argue that more advanced calibration methods \cite{nagel2019data,banner2019post} may improve final accuracy as they are uncoupled to our proposed approach of using synthetic data. Unless otherwise specified, 200 calibration steps are used with batch size 256. Samples are drawn with replacement out of a balanced dataset with size limited to $1\%$ of the training dataset size. Standard data augmentations (e.g., random crop and mirroring) are applied on each input batch to maximize the utilization of available data. Additionally, results are reported as mean and standard deviation with 5 different seeds.

\textbf{Fine-tuning details.} In our KD setting, the float model is the teacher while the student is its quantized variant. Specifically, we apply the most demanding configuration from the previous calibration experiment to create the student. Additionally, teacher predictions are used as targets without additional labels. Furthermore, optimization settings were fixed throughout each set of experiments to enable a fair comparison: all models are optimized with a fixed number of Stochastic Gradient Decent (SGD) iterations, the learning rate scheduler starts with a short warm-up phase followed by a cosine decay phase, while for each optimization step, a batch is drawn with replacement from the target dataset and augmented using standard augmentations methods. We also follow McKinstry et al., \cite{mckinstry2018discovering} and freeze the dynamic range of the student activations during the entire optimization process (we didn't find the suggestion of freezing the dynamic range of the weights necessary).

We also take advantage of the shared teacher and student networks' structure, and apply a simple tweak under the assumption that a good low precision proxy exists within the vicinity of the reference model in the parameters space. Specifically, we use intermediate layer outputs to compute the smoothed-$\ell_1$ distance between the teacher and the student features, we named this loss Intermediate Quantization (IQ) loss.   We found IQ generally lead to a more stable training convergence under extreme quantization and improve KD results for both synthetic and real samples. Additionally, we apply in batch MixUp technique on the inputs similar to \cite{zhang2017mixup}, without mixing the teacher's outputs. The full distillation process is illustrated by algorithm-\ref{alg:dist}, along with an ablation experiment results (table-\ref{tab:ablation}).
\begin{algorithm}[h]\smaller\DontPrintSemicolon
    \Input{$T:$ the teacher model with parameters $\Theta_t$} 
    \Input{$L:= \{l^i_{aux}\}^N$: a set of auxiliary layers (e.g., pointers)}
    \Input{$\mathcal{D}:$ dataset}
    \Param{$criterion_{IQ}, criterion_{KD}$}
    \Param{$\alpha, \beta, \theta_{mix}$}
    \Param{$quantizer(\theta_{q}), optimizer(\theta_{opt})$}  
    \Output{a fine-tuned quantized model}
    
    Init: $S \xleftarrow[]{} quantizer(T,\mathcal{D},\theta_{q})$ \tcp*{\smaller quantize and calibrate the student}
    \For{$X_i \in \mathcal{D}$}
    {
        Compute: $\hat{X}_i \xleftarrow[]{} InputMix(X_i,\theta_{mix})$ \\
        Compute: $logits_t,aux_t \xleftarrow[]{} T.forward(\hat{X}_i,L)$ \\
        Compute: $logits_s,aux_s \xleftarrow[]{} S.forward(\hat{X}_i,L)$ \\
        Compute: $loss_{IQ} \xleftarrow[]{} \sum_{l_i\in L}{criterion_{IQ}(aux_t^{l_i},aux_s^{l_i})}$ \tcp*{\smaller e.g., Smooth-L1 loss}

        Compute: $loss_{KD} \xleftarrow[]{} criterion_{KD}(logits_{t},logits_{s})$ \tcp*{\smaller e.g., KL-Divergence loss}
        Compute: $loss \xleftarrow[]{} \alpha*loss_{KD} + \beta*loss_{IQ}$ \\
        Compute: $G_{S} \xleftarrow[]{} backward(loss,\Theta^{i}_s)$ \tcp*{\smaller i.e., $\frac{\partial loss}{\partial{\Theta^{i}_s}},\ where\ \Theta^{i}_s\ are\ the\ weights\ of\ S\ $}
        Update: $\Theta^{i+1}_s \xleftarrow[]{} optimizer.step(G_{S},\theta_{opt})$ 
    }
\caption{Quantized Distillation with IQ}
\label{alg:dist}
\end{algorithm}
\begin{table}[h]
    \caption{KD ablation study on input-mixing and IQ. We fine-tune a quantized ResNet-44 (2w4a, first \& final layers are in 4 bits) on CIFAR10 dataset with varying dataset size. }
    \small\centering
    \begin{tabular}{r|c|c|c|c}
        \textit{\#samples} & \textit{KD} & \textit{KD+IQ} & \textit{KD+Mix} & \textit{KD+IQ+Mix}\\
        \hline
        1   & 76.32 & 79.67 & 82.49 & 83.68 \\
        10  & 81.12 & 81.71 & 84.01 & 85.44 \\
        100 & 80.16 & 81.64 & 85.05 & 85.95\\
        1000 & 84.03 & 84.66 & 87.2 & 87.28 \\
        2000 & 85.41 & 85.74 & 87.84 & 88.36 \\
        
    \end{tabular}
    \label{tab:ablation}
\end{table}

\textbf{Small scale experimental details.}
We wish to evaluate the applicability of synthetic samples for model calibration and distillation compared to real data. For this experiment we use ResNet44 and Wide-Resnet28-10 \cite{he2016deep,Zagoruyko_2016} on CIFAR10, CIFAR100 datasets \cite{krizhevsky2009learning} respectively.

We start the process by following each of the proposed generation schemes to produce synthetic samples. Next, we apply increasingly demanding quantization configurations to the models and calibrate each variant using samples drawn from the target dataset which is limited to 50 samples per class. After the initial calibration, we optimize the selected models for 16,000 iterations of SGD with a batch size of 512 and 256 for ResNet44 and Wide-ResNet28-10 respectively. IQ loss is computed as the mean of the smoothed-l1 over the outputs of blocks 2-5 from ResNet architecture and multiplied by a scale of 0.001.

\begin{table*}[h]
\centering
\caption{CIFAR validation accuracy for low precision models using data from synthetic and real datasets. The \textit{Settings} column refers to the number of bits used as well as the number of samples-per-class in the dataset. The \textit{Reference} column presents the original training data results. Fine-tune (KD) and calibration with synthetic samples is on par compared to real data, while the results reflect the importance of using statistics preserving samples for demanding compression settings.}
\small\begin{tabular}{l|l|l|l|l|l|l}
            & \multicolumn{1}{c}{\textit{Settings}}   & \multicolumn{1}{c}{\textit{Reference}}   & \multicolumn{1}{c}{$BNS$}         & \multicolumn{1}{c}{$\mathcal{I}$} & \multicolumn{1}{c}{$BNS+\mathcal{I}$}   & \multicolumn{1}{c}{$\mathcal{G}$} \\
\hline
\multicolumn{7}{c}{\textbf{ResNet-44 - CIFAR10, fp32 accuracy 93.23}} \\
\hline
\textit{Calibration} & 4w8a, 50\footnote[1] & 92.18    (0.05)  & 92.21 (0.03)  & 92.37 (0.03)  & 92.25 (0.04)        & 92.24 (0.01) \\
            & 4w4a, 50        & 89.19 (0.15)  & 88.5 (0.13)   & 87.44 (0.13)  & 89.1 (0.09)         & 87.51 (0.18) \\
            & 2w4a, 50\footnote[2]    & 19.47 (0.16)  & 19.14 (0.05)  & 19.42 (0.2)   & 19.49 (0.1)         & 18.48 (0.2) \\
\textit{KD} & 2w4a, 50\footnote[2]  & 90.27 & 88.52 & 79.62   & 88.16  & 70.03  \\
            & 2w4a, 4000\footnote[2]  & 91.24 & 89.6  & 79.59   & 88.51  & 71.02  \\ 
\hline
\multicolumn{7}{c}{\textbf{Wide ResNet-28-10 - CIFAR100, fp32 accuracy 83.69}} \\ 
\hline
\textit{Calibration}   & 8w8a, 50     & 82.96 (0.03)  & 83.11 (0.04)  & 83.01 (0.04)  & 83.06 (0.02)  & 83.15 (0.04) \\
              & 4w4a, 50\footnote[1]     & 76.96 (0.15)  & 74.4 (0.11)   & 62.78 (0.32)  & 74.56 (0.15)  & 58.79 (0.09) \\
              & 4w4a, 50     & 64.52 (0.15)  & 61.48 (0.08) & 51.22 (0.23)  & 61.82 (0.02)  & 47.81 (0.19) \\
\textit{KD}            & 4w4a, 50     &   81.7    & 79.21 &  53.64            & 78.88             & 64.8 \\ 
              & 4w4a, 200    &   82.11   & 79.16 &  54.02            & 78.75             & 65.15 
\end{tabular}%

\small\footnote[1]{}{First \& final layers are in 8 bits}
\small\footnote[2]{}{First \& final layers are in 4 bits}
\label{tab:cifar}
\end{table*}

\textbf{Small scale results.} Results on post-calibration and KD fine-tuning are shown in table-\ref{tab:cifar}. We find that synthetic samples which minimize $BNS$ are superior to samples from other generation schemes, while achieving comparable accuracy to real data, as the fine-tuned models recover from extreme accuracy degradation. Additionally, we observe a surprising outcome regarding the usefulness of Gaussian samples for calibration and distillation --- under mild compression requirements. We noticed that performing performing KD with Gaussian samples requires an additional step to freeze all batch normalization layers, to prevent corruption of the intermediates statistics of the original training data. We believe the full potential Gaussian samples are yet to be discovered and we encourage it as a future research direction. One simple path is truncating the model to disjoint intervals between BN layers then retrain each interval using KD over synthetic data generated by the applying Gaussian scheme on the statistics of the previous BN layer. We also note that Inception related schemes appear to be more sensitive to the hyper-parameters choice (see appendix-\ref{apdx:hp}).

\textbf{Large scale experimental details.} To demonstrate the applicability of our findings in a large-scale setting, we provide our results on ImageNet \cite{russakovsky2015imagenet}. ImageNet has been noted to be a challenging generative task even when full data access is granted, due to the relatively large spatial size of the data and the number of classes \cite{salimans2016improved}. Previous work \cite{nagel2019data} showed that 8-bit models can be calibrated without any data if BN layers exist. However, they did not investigate more challenging numerical precision levels and thus did not need to fine-tune the model. We conduct our experiment with a pre-trained models from \textit{torchvision model-zoo} \cite{paszke2017automatic}. First, we generate 10K/100k samples (see examples in appendix-\ref{apdx:gen_samples}), then perform KD to fine-tune the quantized student, following similar settings to ones described in the previous section, except for a longer regime of 44000 steps and a batch size of 256. Throughout our experiments, the quantization scheme is fixed to a widely used method described in \cite{krishnamoorthi2018quantizing}. Methods that improve this scheme (such as bit allocation, bias correction \cite{banner2019post}, and equalization \cite{nagel2019data,meller2019same}) are orthogonal to this work and are expected to improve accuracy respectively.

\begin{table*}[h]
\centering
\caption{ImageNet validation accuracy for low precision models using data from synthetic and real datasets. We use pre-trained weights from \textit{torchvision model-zoo} \cite{paszke2017automatic} for several meta-architectures under varying compression settings. $BNS$ results are on par with the real data, while KD performs well on the limited size dataset compared to standard cross-entropy loss, see appendix-\ref{apdx:ce_imgnet} for more details.}
\small\begin{tabular}{l|l|l|l|l|l|l}
            & \multicolumn{1}{c}{\textit{Settings}}   & \multicolumn{1}{c}{\textit{Reference}}   & \multicolumn{1}{c}{$BNS$}         & \multicolumn{1}{c}{$\mathcal{I}$} & \multicolumn{1}{c}{$BNS+\mathcal{I}$}   & \multicolumn{1}{c}{$\mathcal{G}$} \\
\hline
\multicolumn{7}{c}{\textbf{ResNet-18 \cite{he2016deep} - ImageNet, fp32 accuracy - 69.75}} \\ 
\hline
\textit{Calibration} & 8w8a, 10\footnote[1] & 69.63 (0.03) & 69.55 (0.01) & 69.6  (0.02) & 69.57 (0.04) & 68.94 (0.02)\\
            & 4w4a, 10\footnote[2] & 54.72 (0.06) & 55.29 (0.1) & 38.25 (0.1)  & 55.49 (0.06) & 53.02 (0.1) \\
\textit{KD} & 4w4a, 10\footnote[2] & 68.63        & 67.98        & 62.8        & 68.06        & 63.98 \\
            & 4w4a, 100\footnote[2]& 68.68        & 68.14        & 63.1        & 67.95        & 63.58 \\
\hline
\multicolumn{7}{c}{\textbf{MobileNet-V2 \cite{Sandler_2018} - ImageNet, fp32 accuracy - 71.88}} \\ 
\hline
\textit{Calibration} & 8w8a, 10\footnote[1] & 71.26 (0.05) & 71.34 (0.03) & 71.2 (0.01) & 71.32 (0.04) & 71.17 (0.02) \\
            & 4w4a, 10\footnote[3] & 15.1  (0.1) & 16.17 (0.1) & 10.55 (0.04) & 16.1  (0.06) & 13.36 (0.04)  \\
\textit{KD} & 4w4a, 10\footnote[3] & 68.5        & 66.4        &   53.13      &    66.07   & 36.95 \\
\hline
\multicolumn{7}{c}{\textbf{DenseNet-121 \cite{Huang_2017} - ImageNet, fp32 accuracy - 74.65}} \\ 
\hline
\textit{Calibration} & 8w8a, 10 & 74.41 (0.01) & 74.41 (0.03) & 74.22 (0.02) & 74.23 (0.02) & 74.16 (0.02)  \\
            & 4w4a, 10 & 45.27 (0.04) & 43.54 (0.15) & 40.46 (0.1) & 43.88 (0.09) & 46.89 (0.03) \\
\textit{KD} & 4w4a, 10 & 71.08 & 71.26         & 63.98     & 70.72 & 63.59
\end{tabular}

\small\footnote[1]{}{Compared to \cite{nagel2019data} (without weights adjustments): 69.6, 69.7}
\small\footnote[2]{}{First \& final layers are in 8 bits}
\small\footnote[3]{}{1x1 convolution layers are in 8 bits}
\label{tab:r18_imagenet}
\end{table*}

\textbf{Large scale results.} Table-\ref{tab:r18_imagenet} presents promising results for recovering lost model accuracy on ImageNet classification task. To the best of the authors' knowledge, this is the first time a compressed model was successfully fine-tuned without using any real data other than the trained model itself at this scale.

We observe an accuracy gap (relative 1.5\% degradation) between standard Cross-Entropy (CE) objective and KD when fine-tuning the model with the full dataset is available. However, results are in favor of KD when using significantly fewer samples (a detailed comparison is available in appendix-\ref{apdx:ce_imgnet}). We believe this can be attributed to a couple of factors. First, semi-supervised KD does not use the ground truth label information. Thus, final accuracy depends solely on the prediction quality of the teacher, whereas label information can be used to penalize the student when repeating similar mistakes made by the teacher. Additionally, we speculate that a given bias in the reference model's prediction towards certain classes may degrade the student accuracy when training on raw teacher outputs.

Finally, we conclude that using synthetic data leads to either equivalent or slightly lower final accuracy compared to real data. We associate the accuracy loss with the synthetic data optimization process (i.e. higher $\mathcal{J}_{KL}$  than real data), as well as with the limited ability of the provided weights to capture the true variance of the data, which in turn leads to a limited sample space.

\begin{figure*}[h]
     \centering
     \begin{subfigure}[t]{0.45\textwidth}
         \centering
         \includegraphics[width=\textwidth]{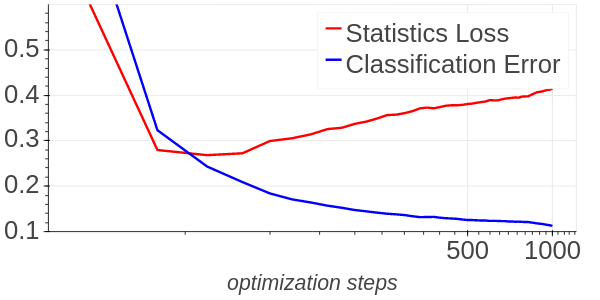}
         \caption{Measured $BNS$ loss quickly diverges when optimizing Inception loss, indicating that the generated samples may be out of the original data distribution.}
         \label{fig:inception_gen}
     \end{subfigure}
     \hfill
     \begin{subfigure}[t]{0.45\textwidth}
         \centering
         \includegraphics[width=\textwidth]{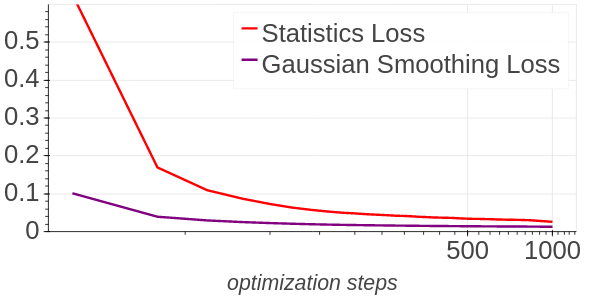}
         \caption{Gaussian smoothing loss is decreasing during $BNS$ loss optimization, revealing a correlation between statistics loss and the spatial structure within feature maps.}
         \label{fig:stats_gen}
     \end{subfigure}
        \smaller\caption[c]{Generating synthetic samples from ResNet-44 trained on CIFAR10.}
        \label{fig:gen_figs}
\end{figure*}
 \subsection{Analysis of Data Generation Schemes}
For this section we consider ResNet44 trained on CIFAR10 as the reference model, in an attempt to understand the relations between generation schemes and provide further explanation to the success of $BNS$ scheme over its counterparts. We will consider the differences between methods from the perspective of the optimization objectives for generating synthetic samples. Additionally, we evaluate the impact of the internal statistics' divergence on KD's potential for recovering lost accuracy.

\textbf{Monitoring Internal Statistics.}
As an initial step in our evaluation, we perform a pair of experiments on Inception and $BNS$ schemes. In each experiment, we follow the process of one scheme and observe the other's objective behavior. More specifically, we generate a batch of samples using the Inception scheme while monitoring the behavior of the internal statistics through $\mathcal{J}_{KL}$. For the alternative view, we generate a second batch of samples using the $BNS$ scheme, and observe the input smoothness loss. In both experiments, we perform 1000 optimization steps on a batch of 128 samples. For each iteration we use in-batch augmentations with N=4, additionally, a Gaussian kernel of size 3x3 and sigma=1 is used for the smoothing loss operation. Results are described in Figure \ref{fig:gen_figs}. 

Under the Inception generation scheme, we find evidence for the divergence of the internal statistics as seen in figure \ref{fig:inception_gen}. Additionally, figure \ref{fig:stats_gen} indicates that minimizing $\mathcal{J}_{KL}$ leads to lower smoothness loss. Interestingly, figure \ref{fig:inception_gen} the initial improvement in $\mathcal{J}_{KL}$ during optimization of $\mathcal{I}$ hints to the existence of a connection in a reversed direction as well. However, smoothing loss alone proves to be insufficient in regularizing the internal statistics' divergence as the Inception loss greedily attempts to maximize the target class predictions and the measured $\mathcal{J}_{KL}$ is increasing after several iterations.

\textbf{BN-Stats and Model Accuracy.} For the next part of our evaluation, we generated samples from the reference model using the $BNS$ scheme. During the sample generation phase, several snapshots of the training data were saved at different step intervals. The snapshots reflect different stages of the $\mathcal{J}_{KL}$ loss curve. We then perform a series of KD experiments on the quantized reference model to observe the impact of dataset size and $\mathcal{J}_{KL}$ loss value on the final model accuracy. 
Each experiment is repeated with five different seeds, the student model precision configuration is fixed to 4-bit activations and 2-bit weights except for the first and last layers which are using 4 bit.
\begin{figure*}[h]
     \centering
     \begin{subfigure}[t]{0.48\textwidth}
         \centering
         \includegraphics[width=\textwidth]{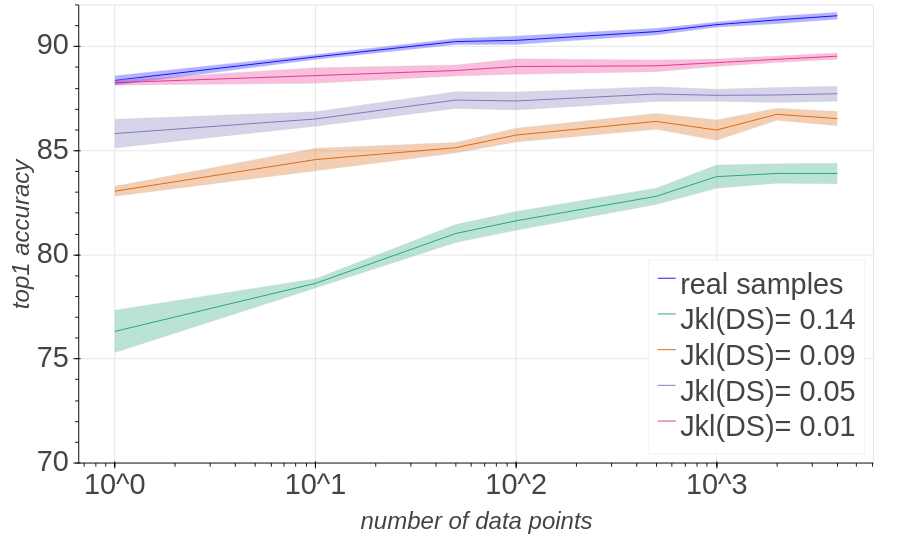}
         \caption{Dataset size impact on final model accuracy.}
         \label{fig:dist_dataset_size_acc}
     \end{subfigure}
     \hfill
     \begin{subfigure}[t]{0.48\textwidth}
         \centering
         \includegraphics[width=\textwidth]{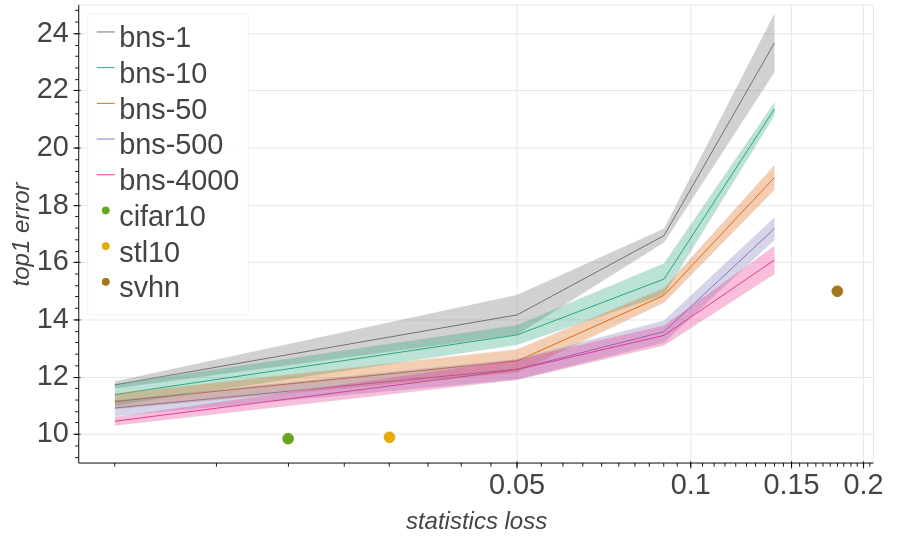}
         \caption{Statistics loss impact on final error.}
         \label{fig:dist_stat_loss_acc}
     \end{subfigure}
        \caption{KD on a quantized ResNet-44 model (2w4a, CIFAR10) using real and $BNS$ data with varying number of data points (per-class). Demonstrates the connection between dataset size and $\mathcal{J}_{KL}$ with model accuracy. Providing more synthetic samples improves final accuracy.}
        \label{fig:dist_figs}
\end{figure*}

Final results are presented in Figure \ref{fig:dist_figs} as follows: Figure \ref{fig:dist_stat_loss_acc} demonstrates that samples with lower statistic loss lead to better validation accuracy, while Figure \ref{fig:dist_dataset_size_acc} shows results are very close to real data with slight degradation under identical settings.

\subsection{Data Correspondence: BNS Across Datasets}
In this section we observe how the internal statistics of a model respond to different inputs, illustrating that similar input data should not lead to significantly different internal activation statistics.
In table-\ref{table:ds_stats} we demonstrate this by measuring $\mathcal{J}_{KL}$ of various datasets on a set of pre-trained ResNet44 \cite{he2016deep} models. We also measure the $\mathcal{J}_{KL}$ divergence in response to the original training data modified with Fast Gradient Sign Method (FGSM \cite{goodfellow2014explaining}). A small perturbation ratio of $\epsilon=0.1$ is used to reduce model accuracy without considerably altering the perceived input.
\begin{table*}[h]
    \centering
    \caption{The value of $\mathcal{J}_{KL}$ normalized by the value of $\mathcal{J}_{KL}$ that is measured on the original training dataset. Values are computed on the entire train split, raw measurements are available in appendix-\ref{apdx:bns_measures}.}
    \small\begin{tabular}{l|c|c|c|c|c|c|c}
     \textit{Train/Measure} & \textit{CIFAR10}  &\textit{CIFAR100}&   \textit{MNIST}& \textit{SVHN} & \textit{STL10} & \textit{Random}&  \textit{FGSM\footnote[1]{}}  \\ 
    \hline
    \textit{CIFAR10}  &    1.0     &    1.0&    6.6&    7.9        &1.3&    21.7&    3.3\\
    \textit{CIFAR100} &    1.1     &    1.0&    3.9&    4.1&    1.3&    14.8&    1.9\\
    \textit{MNIST}    &    103.5   &    111.5&    1.0&    196.0&    138.5&    327.5&    2.0\\
    \textit{SVHN}     &    1.4     &    1.2&    1.2&    1.0&    2.7&    4.1&    0.9\\
    \end{tabular}

    \smaller\footnote[1]{}{Ratio grows with the $\epsilon$ parameter of FGSM, as the image perturbation becomes more apparent.}
    \label{table:ds_stats}
\end{table*}
As expected from a similarity score, table-\ref{table:ds_stats} demonstrates that $\mathcal{J}_{KL}$ maintains its proportion compared to the original training dataset, although magnitude may change across models and datasets. We further observe that in some cases, perturbed samples results in a significant increase in the internal statistics' divergence. This invites further exploration in the direction of possible applications of monitoring internal statistics to explain and detect adversarial attacks through tailored outlier sensitive measures.

\section{Discussion and Future Work}
In this work, we addressed the case of data-free KD for model quantization under realistic data and hardware acceleration constrains. To overcome this challenge, we suggested a set of new sample-generation methods, and demonstrated the applicability of such samples for a true data-free model compression in both small and large scale image classification tasks. In the process we explored techniques to leverage such samples in a student-teacher training procedure.

In addition, our best performing method, $BNS$ leveraged the common batch-norm layer to generate examples that mimic the per-channel mean and variance of each BN layer input, by explicitly optimizing $\mathcal{J}_{KL}$ divergence loss. 
We show evidence that optimizing this loss, results in smooth input features, indicating a strong connection between BN-Stats loss and the local structure of the input (figure-\ref{fig:stats_gen}). This invites further study to determine the viability of the method for the reproduction of inputs other than images, where prior knowledge may be harder to apply directly. 

Alternatively, our random data approach $\mathcal{G}$ can serve as a cheap replacement to the real data. In particular, for calibration under mild quantization (8-bit) demands. We also noticed that compared to $BNS$, the naive inception scheme $\mathcal{I}$ generates samples with a high model prediction confidence, yet causes the internal statistics' divergence to grow significantly (figure-\ref{fig:inception_gen}). Furthermore, the success of $BNS$, leads us to consider it as a possible measure for evaluating correspondence between the reference dataset, used during model's training, and alternative samples.

To the best of our knowledge, this is the first time a data-free distillation-based approach was applied to a model and achieved final accuracy results comparable with real samples --- under identical optimization conditions and at large scale. Ultimately, we believe our approach can open a path to improve data privacy by reducing the extent of real data exposure during the production phase of deployment.

The release of this paper was followed by \cite{yin2019dreaming,cai2020zeroq}, which utilize the same principle of leveraging the internal statistics' divergence for a range of data-free tasks, including low precision calibration and knowledge distillation. These papers explore interesting ideas such as adaptive sample generation and layer sensitivity heuristics for mixed-precision, while their results further support our findings.
In contrast to \cite{yin2019dreaming}, which presents a wide range of experiments on data-free applications such as classical KD and pruning with a limited analysis, we focused our attention on the applicability of the method for DNN's deployment on low precision accelerators and  provide further insight into the success of the method and to the importance of the internal divergence measure. Additionally, we present data-free KD results for low precision models, which are not explored by \cite{cai2020zeroq}, and show highly compressed models can recover from extreme accuracy degradation which enables much more demanding configurations without requiring access to the original training data.

We consider two drawbacks of the proposed method. One is the computational cost of generating samples through back-propagation, which can impede the practical use with large scale models for continuous train-deploy scenarios. However, we note that as long as the new training data does not significantly change, the behavior of the internal statistics of the generated samples can potentially be shared --- to avoid reproducing an entirely new synthetic dataset at each deployment cycle. We leave the exploration of cross-model and cross-dataset applications for future work. Second, we find the $BNS$ method produces datasets which are unbalanced in terms of the mean output distribution of the reference model, due to the lack of explicit conditioning on the model output. However, our experiments with $BNS+\mathcal{I}$ did not show a dramatic improvement despite their added balance control and the additional information injection from the final layer weights. We suspect this is due to the tension between the $BNS$ objective and $\mathcal{I}$ which may require balancing or a longer optimization to reach comparable $\mathcal{J_{KL}}$. Still, there is a lot to unveil and we are excited by the diverse opportunities to exploit the suggested scheme beyond model compression. For instance, applying $BNS$ measure to detect outliers and adversarial examples, or to avoid \textit{catastrophic forgetting} in a continual learning setting.

{\small
\bibliographystyle{CVPR2020/ieee_fullname}
\bibliography{egbib}
}
\onecolumn
\clearpage
\begin{appendices}


\section{Hyper-Parameters}
\label{apdx:hp}
\subsection{Generation}
Using ADAM optimizer with learning rate 0.1, betas(0.9,0.999) for all experiments, when image-prior scale is non-zero we use a $\mathcal{N}$(0,1) smoothing kernel of size 5. Generally batch size is chosen to fit in available memory (statistics are accumulated across devices).
\begin{table}[h]
    \small\centering
    \begin{tabular}{l|c|c|c|c|c}
    \textit{scheme}   & \textit{batch size} & \textit{in-batch} & \textit{steps} & \textit{lr drops (1e-1) }& \textit{scales (bns,cls,prior)}\\
    \hline
    \multicolumn{6}{c}{\textbf{ResNet-44-CIFAR10}}\\
    \hline
    $BNS$ & \multirow{3}{*}{800}& \multirow{3}{*}{4} &   \multirow{3}{*}{1000}  & \multirow{3}{*}{800} &    1,0,0  \\
    $BNS+\mathcal{I}$ & & & &   &  1,1e-3,1\\
    $\mathcal{I}$ & & & &       &  0,1e-3,1\\
    \hline
    \multicolumn{6}{c}{\textbf{Wide ResNet-28-10-CIFAR100}} \\
    \hline
    $BNS$ & \multirow{3}{*}{256}& \multirow{3}{*}{4} &   \multirow{3}{*}{1000}  & \multirow{3}{*}{800} &    1,0,0  \\
    $BNS+\mathcal{I}$ & & & &   &  1,1e-3,1\\
    $\mathcal{I}$ & & & &       &  0,1e-3,1\\
    \hline
    \multicolumn{6}{c}{\textbf{ResNet-18-ImageNet}} \\
    \hline
    $BNS$ & \multirow{3}{*}{80}& \multirow{3}{*}{4} &   \multirow{3}{*}{1000}  & \multirow{3}{*}{800} &    1,0,0  \\
    $BNS+\mathcal{I}$ & & & &   &  1,1e-3,1\\
    $\mathcal{I}$ & & & &       &  0,1e-3,1\\
    \hline
    \multicolumn{6}{c}{\textbf{MobileNet-V2-ImageNet}} \\
    \hline
    $BNS$ & \multirow{3}{*}{400}& \multirow{3}{*}{2} &   6000 & 1500,4200 &  1,0,0  \\
    $BNS+\mathcal{I}$& &                             &  10000 & 1500,6000 &  1,1e-4,0\\
    $\mathcal{I}$ & &                                & 1000   & 800       &  0,1e-4,1\\
    \hline
    \multicolumn{6}{c}{\textbf{DenseNet-121-ImageNet}}\\ 
    \hline
    $BNS$ & \multirow{3}{*}{200}& \multirow{3}{*}{2} & 6000  & 1500,4200 & 1,0,0  \\
    $BNS+\mathcal{I}$ &         &                    & 10000 & 1500,6000 & 1,1e-4,0\\
    $\mathcal{I}$     &         &                    & 1000  & 800       & 0,1e-4,1
    \end{tabular}
\end{table}

\subsection{Hyper-Parameters sensitivity for calibration}
Here we present evidence of hyper-parameter sensitivity for Inception related generation schemes. Specifically, we demonstrate how changing the variance parameter of the Gaussian smoothing kernel, used as the image prior, to regularize the optimization process, impacts the usability of the generated data. In table-\ref{tab:sensitivity_}, we compare calibration results for quantized ResNet44 from \textit{Small scale results} section.
\begin{table*}[h]
        \caption{Post calibration validation accuracy, additional fixed hyper-parameters: kernel 5x5, inception loss scale 0.001. Results on $\mathcal{I}$ and $BNS+\mathcal{I}$ datasets appear to be sensitive to hyper-parameter adjustments. We report our best results without performing an exhaustive search.}
    \small\centering
    \begin{tabular}{l | c | c}
        sigma &  $\mathcal{I}$ & $BNS+\mathcal{I}$ \\
        \hline
        \multicolumn{3}{c}{\textbf{ResNet-44, 4w4a, real data calibration accuracy - 89.19 (0.15)}} \\
        \hline
         0.375  & 89.22 (0.22) & 88.87 (0.22)\\
         1.0   &  87.44 (0.13) & 89.1 (0.09) \\
        \hline
        \multicolumn{3}{c}{\textbf{ResNet-44, 4w4a\footnote[1], real data calibration accuracy - 91.64 (0.13)} }\\
        \hline
         0.375& 91.26 (0.16) & 91.3 (0.1) \\
         1.0 & 90.89 (0.11)& 91.76 (0.22) \\
        \hline
        \multicolumn{3}{c}{\textbf{ResNet-44, 4w8a\footnote[1]], real data calibration accuracy - 92.18 (0.05)}}\\
        \hline
        0.375   & 92.04 (0.03) & 92.33 (0.04)  \\
       1.0  & 92.37 (0.03) & 92.25 (0.04)\\
    \end{tabular}

    \footnote[1]{}{First \& final layers are in 8 bits}
    \label{tab:sensitivity_}
\end{table*}

\subsection{Distillation}
Algorithm-\ref{alg:dist} describes the general KD framework with IQ and input-mixup tweaks (see ablation in table-\ref{tab:ablation}). These are not strictly required when the full dataset is available, yet tend to offer improved convergence when the available dataset is limited in size. We use standard KL divergence as KD loss, smoothed-l1 is used for IQ term and the calibration algorithm is as described in section-\ref{sec:calib}. We use SGD with a learning rate of 0.1 and a batch size of 256 in all experiments, while sampling with replacement for a fixed number of steps per epoch (CIFAR-200 ,ImageNet-400), additional shared hyper-parameters: $\alpha, \beta, \theta_{mixup}$ = 1, 0.01, \{mix\_rate=0.5\}.

\section{Additional results for ImageNet dataset}
\label{apdx:ce_imgnet}
\begin{table}[t]
    \caption{Comparison of dataset size and fine-tune objective impact on ImageNet validation accuracy for ResNet-18 quantized to 4 bits. $^*$ Regime follows \cite{mckinstry2018discovering}.
    }
    \small\centering
    \begin{tabular}{l|l|l|c|c|c}
        \textit{dataset} & \textit{samples} & \textit{objective}  & \textit{steps} & \textit{top1} &  $\mathcal{J}_{KL}$ \\
        \hline
        ImageNet &~1.3M & CE & *550440 &  69.95  & 0.05\\
             &~1.3M & CE & *45036  &  68.36  & \\
             &100K & CE & 44000   &  67.47  & \\
        \hline
        ImageNet     &~1.3M & KLD & 550440  &  68.87  & 0.05\\
             &100K & KLD & 44000   &  68.68  & \\
        \hline
        $BNS$ & 100K & KLD  & 44000  & 68.14  & 0.052\\
        \hline
        $BNS+\mathcal{I}$ & 100K  & KLD & 44000  & 67.95  & 0.059 \\
        \hline
        $BNS$+$BNS+\mathcal{I}$ & 50K+50K  & KLD & 44000  & 68.07  & 0.053 \\
    \end{tabular}

    \label{tab:imgnt_ce_kd}
\end{table}

There exists an accuracy gap between the fine-tuned model and the distilled variants even when the full data-set is available as seen in table-\ref{tab:imgnt_ce_kd}. We believe this can be attributed to using KD without an additional ground truth loss term which is common in an unsupervised setting. Thus, final accuracy depends solely on the prediction quality of the teacher. Whereas additional label information can be used to penalize the student when repeating similar mistakes made by the teacher and contribute to improved final accuracy. Additionally, we speculate that a given bias in the reference model's prediction towards certain classes (see figure-\ref{fig:bns+bns_i_bias}) may degrade the student accuracy when training on raw teacher outputs.

\section{Generating CIFAR10 and ImageNet}
\label{apdx:gen_samples}
In figure-\ref{fig:dist_figs}, we provide a several synthetic samples per class generated from a CIFAR10 trained ResNet44 model, using Inception scheme ($\mathcal{I}$) and samples generated using the BN-stats + Inception scheme ($BNS+\mathcal{I}$). The generation process follows the settings detailed in \textit{Generating data samples} section. Samples seem to share similar visual features between generation schemes. while $BNS+\mathcal{I}$ samples are smoother and appear clearer.

\begin{figure*}[h]
     \centering
     \begin{subfigure}[t]{0.48\textwidth}
         \centering
         \includegraphics[width=\textwidth]{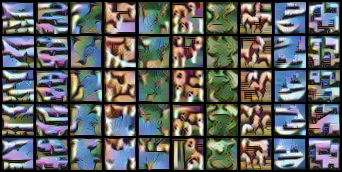}
         \caption{samples generated with $BNS + \mathcal{I}$, Gaussian smoothing Kernel 5x5, sigma 1.0}
     \end{subfigure}
     \hfill
     \begin{subfigure}[t]{0.48\textwidth}
         \centering
         \includegraphics[width=\textwidth]{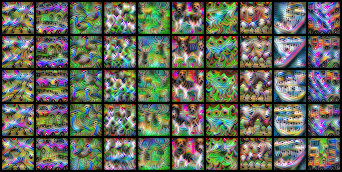}
         \caption{samples generated with Inception loss + Gaussian smoothing Kernel 5x5, sigma 1.0}
     \end{subfigure}
        \smaller\caption{CIFAR10 samples generated from ResNet-44}
        \label{fig:dist_figs}
        
\end{figure*}
Additionally, a subset of samples generated using ResNet-18 model trained over ImageNet dataset is presented in figure-\ref{fig:imagenet_gen}. Visual inspection of those generated samples indicated considerable improvement in detail and diversity over naive inception generation scheme. Furthermore, some instances seem to preserve the represented class physical structure better than others. Despite the lack of consensus regarding objective image quality evaluation methods, we find it is intriguing to further explore means to improve reproduced class visual quality and investigate its connection to the model's prediction quality. We believe the $BNS+\mathcal{I}$ method may serve as a powerful tool for DNNs interpretability, providing insight into the featured learned by DNNs.

\begin{figure}[ht]
    \caption{Comparison of ImageNet and synthetic samples generated from ResNet-18}
    \vspace{-5ex}
    \begin{subfigure}[h]{\textwidth}
      \centering\includegraphics[height=16cm,angle=270]{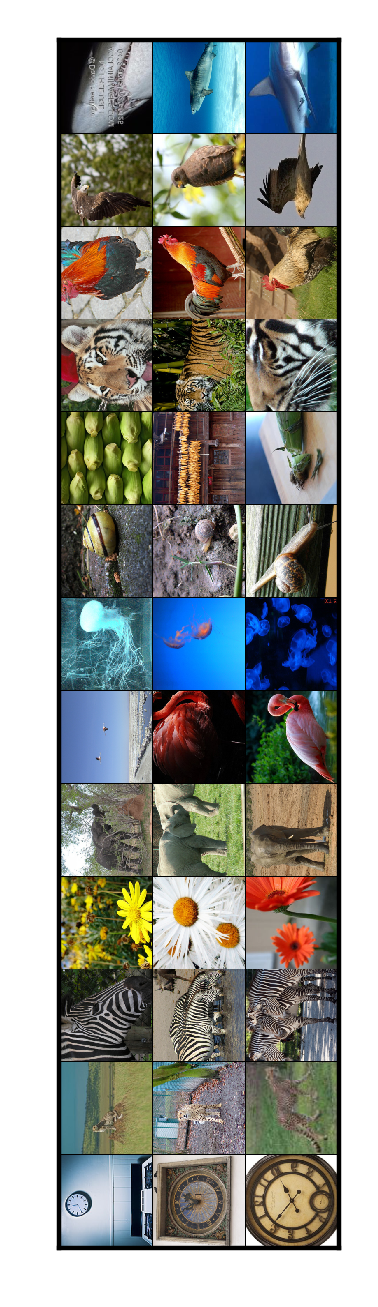}%
      \caption{Reference ImageNet samples}
    \end{subfigure}
    
    \vspace{-5ex}
    \begin{subfigure}[h]{\textwidth}
      \centering\includegraphics[height=16cm,angle=270]{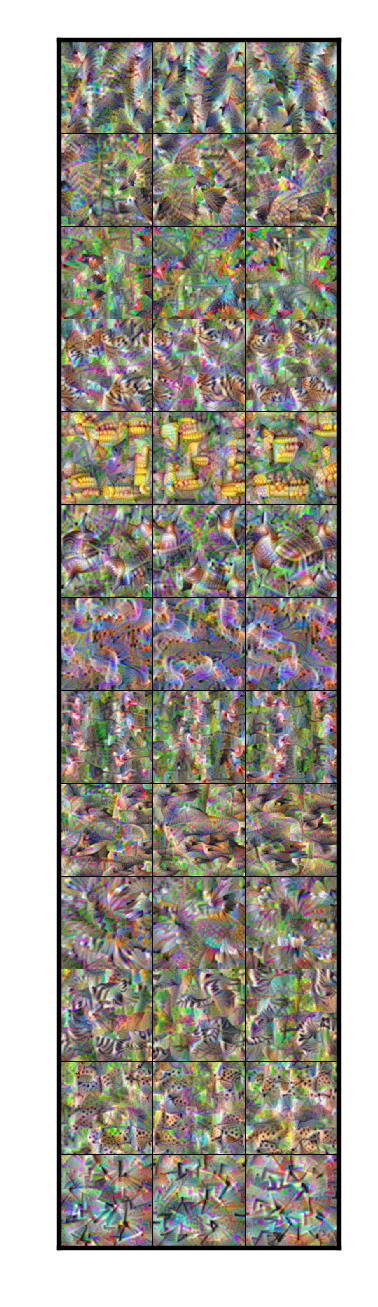}%
      \caption{$\mathcal{I}$, smoothing kernel 9x9, sigma 0.7 yielded best visual result}
    \end{subfigure}
    
    \vspace{-5ex}
    \begin{subfigure}[h]{\textwidth}
      \centering\includegraphics[height=16cm,angle=270]{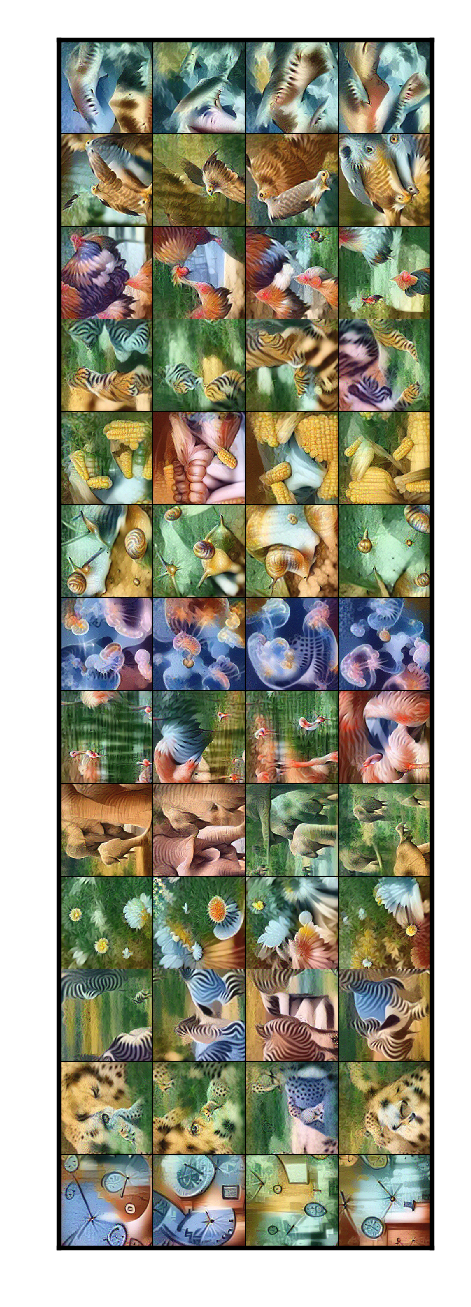}%
      \caption{$BNS+\mathcal{I}$, smoothing kernel 5x5, sigma 1}
    \end{subfigure}
    
    \vspace{-5ex}
    \begin{subfigure}[h]{\textwidth}
      \centering\includegraphics[height=16cm,angle=270]{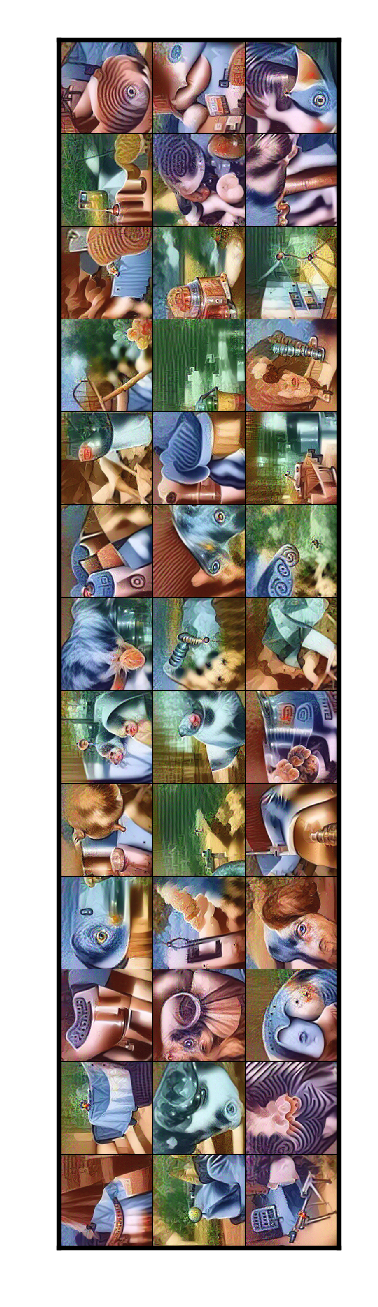}%
      \caption{$BNS$ only}
    \end{subfigure}
\label{fig:imagenet_gen}
\end{figure}

\begin{figure}[ht]
    \centering
    \caption{Additional synthetic samples generated from DenseNet-121}
    \begin{subfigure}[h]{\textwidth}
      \centering\includegraphics[width=0.85\textwidth]{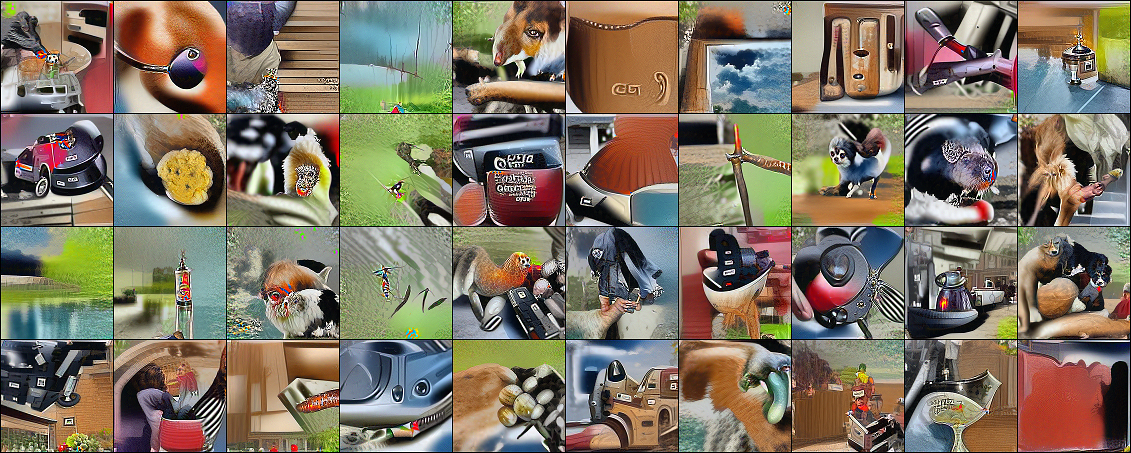}%
      \caption{$BNS$ only samples}
    \end{subfigure}
    \begin{subfigure}[h]{\textwidth}
      \centering\includegraphics[width=0.85\textwidth]{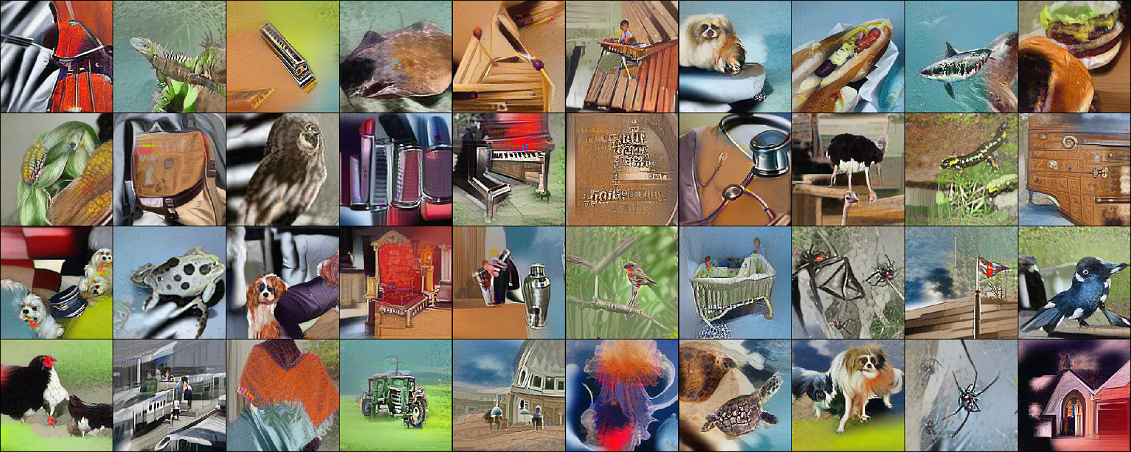}%
      \caption{$BNS+\mathcal{I}$, no image prior is used during sample generation.}
    \end{subfigure}
\label{fig:imagenet_gen}
\end{figure}
\section{BN-Stats measurement}
\label{apdx:bns_measures}
In table-\ref{table:ds_stats_raw}, we provide raw measurements of $\mathcal{J}_{KL}$ on select datasets using pre-trained ResNet44 from table-\ref{table:ds_stats} of the main paper.

\begin{table}[h]
    \caption{$\mathcal{J}_{KL}$ values are computed on the entire training data-set and include all BN layers within the reference model.\\
    $^{*}$Fast Gradient Sign Method (FGSM) was used to create adversarial samples with small perturbation ratio $\epsilon=0.1$, the measured loss generally grows with epsilon.}
    \label{table:ds_stats_raw}
    \small\centering
    \begin{tabular}{l|c|c|c|c|c|c|c}
     Train/Measure &CIFAR10  &CIFAR100&   MNIST& SVHN & STL10 & Random&  $^{*}$FGSM  \\
    \hline
 \textit{CIFAR10}   &    0.023&    0.022&    0.151&    0.182&    0.031&    0.498&    0.075\\
 \textit{CIFAR100}  &    0.065&    0.057&    0.223&    0.236&    0.072&    0.845&    0.106\\
 \textit{MNIST}     &   0.207&    0.223&    0.002&    0.392&    0.277&    0.655&    0.004\\
 \textit{SVHN}      &    0.037&    0.032&    0.032&    0.027&    0.073&    0.111&    0.024
    \end{tabular}

\end{table}
\section{Perceived dataset bias}
\label{apdx:ds_bias}
\begin{figure*}[ht]
        \begin{subfigure}[t]{0.485\textwidth}
            \centering\includegraphics[width=\textwidth]{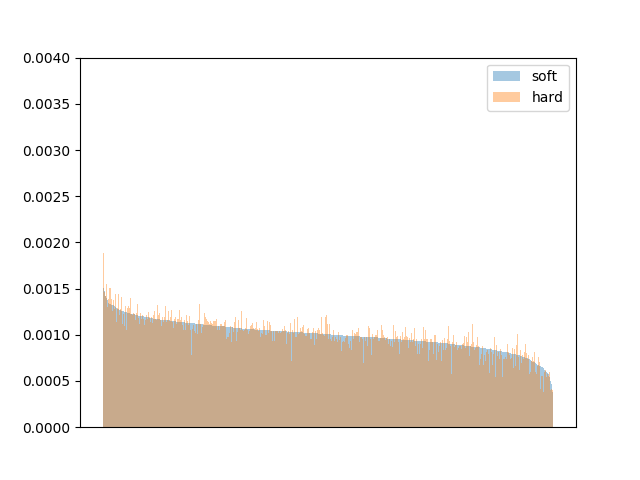}
            \caption[]%
            {\small Reference model (fp32) mean prediction over ImageNet validation set reveals a classifier bias.
        \label{fig:imagenet_bias}}  
        \end{subfigure}
        \hfill
        \begin{subfigure}[t]{0.485\textwidth}  
            \centering\includegraphics[width=\textwidth]{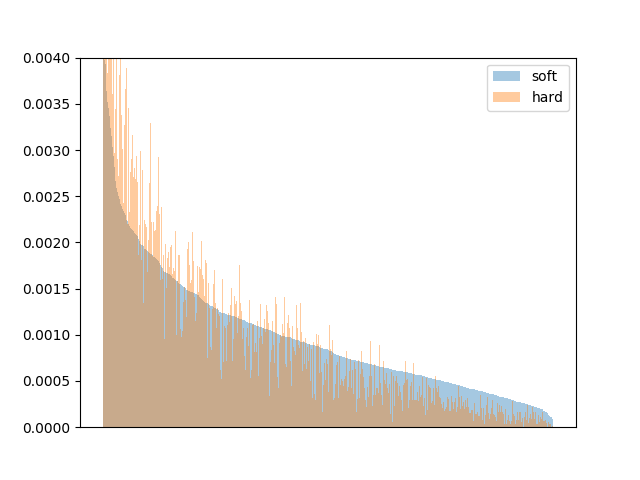}
            \caption[]%
            {\small Reference model (fp32) mean prediction over $BNS$ samples, presents a highly biased preference towards certain classes.}
            \label{fig:bns_bias}
        \end{subfigure}
        \vskip\baselineskip
        \begin{subfigure}[t]{0.485\textwidth}   
            \centering 
            \includegraphics[width=\textwidth]{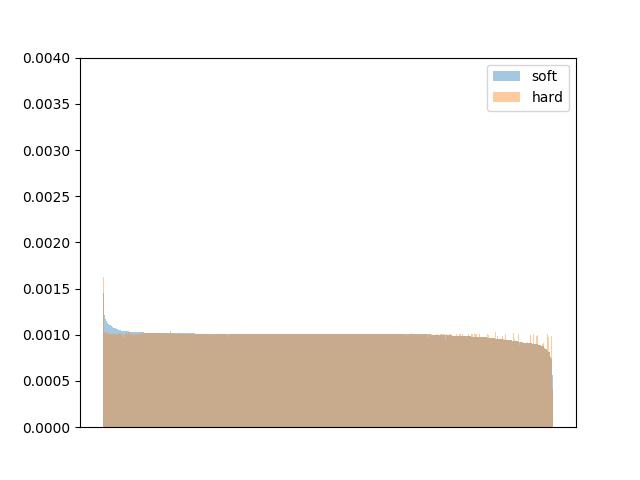}
            \caption[]%
            {\small Reference model (fp32) mean prediction over $BNS+\mathcal{I}$ samples. As expected prediction mean is balanced since the samples are tailored through optimization to favor a single class with uniform target class sampling.}
            \label{fig:bns_i_bias}
        \end{subfigure}
        \quad
        \begin{subfigure}[t]{0.485\textwidth}   
            \centering           \includegraphics[width=\textwidth]{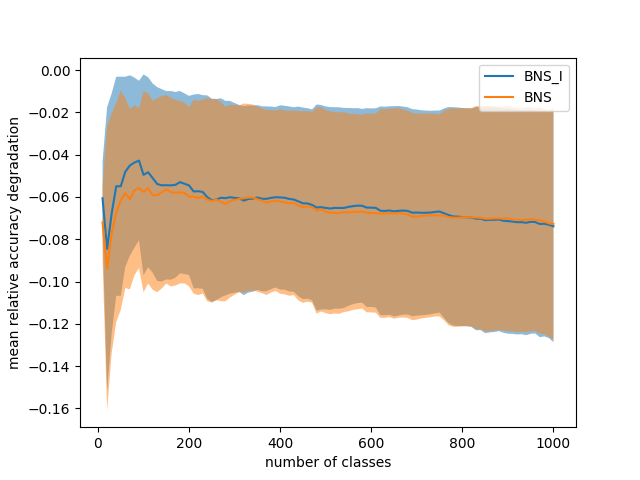}
            \caption[]%
            {\small Mean relative tail degradation on ImageNet validation set. Classes are sorted according to the hard prediction rate from \ref{fig:bns_bias}.
            Degradation ratio is computed as the mean of $\frac{s^{i}_{fp32}-s^{i}_{ft}}{s^{i}_{fp32}}$ over increasing number of classes from least favored to most favored. $s$ denotes per-class accuracy and $i$ is the class index. The $BNS+\mathcal{I}$ fine-tuned model present a small improvement over the $BNS$ variation with respect to the worst case accuracy degradation.}
            \label{fig:degradation}
        \end{subfigure}
\smaller\caption{Bias analysis on ResNet-18, soft and hard plots refer to means of raw model outputs and a hard class index choice (i.e., a hot-1 vector) over the entire validation set}
\label{fig:bns+bns_i_bias}
\end{figure*}

Given a reference classification model, we loosely consider the per-class mean prediction of the model as unbiased if it is close to uniform when presented with a balanced set of examples, since no particular class should be favoured over other classes by an unbiased model. To provide a qualitative evaluation, we record the softmax output of a pretrained ResNet-18 model from \textit{torchvision}, over the entire ImageNet validation set. Then, compute the mean prediction of the "soft" outputs (i.e., $\frac{1}{N}\sum_{i=1}^N{output_i}$) where $output_i\in R^{ \# classes}$ is the model softmax output for sample $i$, as well as the "hard" mean prediction (i.e., $\frac{1}{N}\sum_{i=1}^N e^{argmax\{output_i\}}$ where $e$ is the standard vector). In figures-\ref{fig:imagenet_bias},\ref{fig:bns_bias},\ref{fig:bns_i_bias}, we plot the mean soft and hard predictions of the model for a given dataset, while classes are sorted according to the mean soft prediction. Figure-\ref{fig:imagenet_bias} reveals that although the validation dataset is balanced (ignoring possible annotation errors and class similarity), the model produces a somewhat biased prediction. We hypothesize this bias may impede KD performance as discussed in \textit{Large scale experiment} section, even when the entire dataset is available.

Additionally, in figures-\ref{fig:bns_bias},\ref{fig:bns_i_bias} we provide the mean prediction plots for $BNS$ and $BNS+\mathcal{I}$ datasets appropriately, each containing 100K synthetic samples. Figure-\ref{fig:bns_bias} shows a clear preference towards certain classes, which does not necessarily align with the model's prediction bias over the validation set (figure-\ref{fig:imagenet_bias}). While figure-\ref{fig:bns_i_bias} shows that $BNS+\mathcal{I}$ dataset appears balanced in terms of the mean prediction, which is not a surprising result. However, our KD experiments did not show any significant benefit to using the $BNS+\mathcal{I}$ dataset compared to the $BNS$ dataset in terms of the final validation accuracy, despite the perceived class imbalance.
\section{Mean tail degradation}
\label{apdx:undr_rep}
To further investigate the impact of $BNS$ dataset bias on fine-tuned (quantized) model's accuracy, we consider the mean relative accuracy degradation over N least frequent classes (mean tail degradation). First, we measure the per-class validation accuracy for each of the fine-tuned models (i.e., ResNet18 fine-tuned with $BNS$ or $BNS+\mathcal{I}$ dataset, see \textit{Large scale experiment} section). Then, we compute the relative degradation compared to the float model for each class. Finally, classes are sorted according to the reference model (fp32) mean hard prediction over $BNS$ dataset (figure-\ref{fig:bns_bias}). Essentially, the least frequently predicted classes are considered first under the tail limit (N). Figure-\ref{fig:degradation} shows the mean degradation with an increasing number of classes N, while only considering classes where accuracy degradation is observed (i.e., we ignore classes which present an improvement in accuracy). Figure-\ref{fig:degradation} presents evidence for an improved mean tail degradation, which can serve as motivation for using $BNS+\mathcal{I}$ over $BNS$ dataset since the worst-case accuracy is improved despite the overall accuracy being slightly worse.
\end{appendices}

\end{document}